\documentclass[10pt, a4paper]{article}

\usepackage{lrec-coling2024} % this is the new style

\usepackage[utf8]{inputenc}

\usepackage{cjhebrew}
\usepackage{coptic}
\usepackage{booktabs}
\usepackage{graphicx}
\usepackage{pdfpages}

\usepackage{listings} 
\usepackage{tabularx}

\usepackage{CJKutf8}

\usepackage{comment}
\usepackage{xcolor}
\definecolor{delim}{RGB}{20,105,176}
\definecolor{expletive}{HTML}{7880d7}
\definecolor{copula}{HTML}{59af69}
\definecolor{pivot}{HTML}{cc62a6}
\definecolor{coda}{HTML}{c7893c}
\definecolor{nsubj}{HTML}{673286}
\definecolor{obj}{HTML}{77081b}
\definecolor{obl:lmod}{HTML}{058eca}

\lstdefinelanguage{grew}{
    basicstyle=\scriptsize\ttfamily,%\bfseries,
    numberstyle=\scriptsize,
    extendedchars=true,
    numbersep=8pt,
    showstringspaces=true,
    literate=
     *{EXP}{{{\color{expletive}{\textbf{EXP}}}}}{3}
      {PRO}{{{\color{proform}{\textbf{PRO}}}}}{3}
      {PRED}{{{\color{copula}{\textbf{PRED}}}}}{4}
      {PIV}{{{\color{pivot}{\textbf{PIV}}}}}{3}
      {COD}{{{\color{coda}{\textbf{COD}}}}}{3}
      {nsubj}{{{\color{nsubj}{\textbf{nsubj}}}}}{4}
      {obj}{{{\color{obj}{\textbf{obj}}}}}{3}
      {obl:lmod}{{{\color{obl:lmod}{\textbf{obl:lmod}}}}}{7}
      {pattern}{{{\textbf{pattern}}}}{7}
      {without}{{{\textbf{without}}}}{7}
      {\{}{{{\color{delim}{\{}}}}{1}
      {\}}{{{\color{delim}{\}}}}}{1}
      {[}{{{\color{delim}{[}}}}{1}
      {]}{{{\color{delim}{]}}}}{1},
}

\usepackage{nert} % Nathan's lab macros

\newcommand{\deprel}[1]{\texttt{#1}}
\renewcommand{\lex}[1]{\textbf{\textit{#1}}}
\newcolumntype{Y}{>{\centering\arraybackslash}X}

\newcommand{\affil}[1]{$^\textnormal{\textcolor{gray}{#1}}$}

\title{UCxn: Typologically Informed Annotation of Constructions Atop Universal~Dependencies}

\name{\begin{tabular}{c}Leonie Weissweiler,\affil{1} Nina Böbel,\affil{2} Kirian Guiller,\affil{3} Santiago Herrera,\affil{3} \\Wesley Scivetti,\affil{4} Arthur Lorenzi,\affil{5} Nurit Melnik,\affil{6} Archna Bhatia,\affil{7} \\Hinrich Sch\"{u}tze,\affil{1}  Lori Levin,\affil{8} Amir Zeldes,\affil{4} Joakim Nivre,\affil{9} William Croft,\affil{10} \\Nathan Schneider\affil{4}\end{tabular}
} 

\address{\affil{1}LMU Munich \& MCML, \affil{2}HHU D\"{u}sseldorf, \affil{10}University of New Mexico, \affil{8}Carnegie Mellon University\\ \affil{3}Université Paris Nanterre, CNRS, \affil{4}Georgetown University, \affil{5}Federal University of Juiz de Fora\\ \affil{6}The Open University of Israel, \affil{7}Institute for Human and  Machine Cognition, \affil{9}Uppsala Univ.~and RISE\\
         \texttt{weissweiler@cis.lmu.de, nathan.schneider@georgetown.edu}\\}

\abstract{ 
The Universal Dependencies (UD) project has created an invaluable collection of treebanks with contributions in over 140 languages.
However, the UD annotations
do not tell the full story. Grammatical constructions that convey meaning through a particular combination of several morphosyntactic elements---for example, interrogative sentences with special markers and/or word orders---are not labeled holistically.
We argue for (i) augmenting UD annotations with a ``UCxn'' annotation layer for such meaning-bearing grammatical constructions, and
(ii) approaching this in a typologically informed way so that morphosyntactic strategies can be compared across languages. 
As a case study, we consider five construction families
in ten languages,
identifying instances of each construction in UD treebanks through the use of morphosyntactic patterns.
In addition to findings regarding these particular constructions, our study yields important insights on methodology for describing and identifying constructions in language-general and language-particular ways, and lays the foundation for future constructional enrichment of UD treebanks.
 \\ \newline \Keywords{grammatical constructions, treebanks, Universal Dependencies, typology, corpus annotation}
}

\usepackage{fancyhdr}
\begin{document}
\draftversion{\thispagestyle{fancy}}

\maketitleabstract

\section{Introduction}\label{sec:intro}

The notion of a \emph{construction} is an important concept in grammar as it allows for an analysis of patterns of form and function within languages as well as systematic comparisons across languages. 
Consider the WH-interrogatives in English and Coptic. 
While English uses a combination of WH-words and word order to encode such questions, Coptic typically leaves WH-words in situ, meaning they occur in the same position as non-interrogative pronouns:\footnote{In many cases, prosody or punctuation can also indicate a clause is interrogative. Coptic texts, however, do not use question marks, and e.g.,~web data contains nonstandard punctuation use \cite{SanguinettiEtAl2022}.}

\exg. 
    e- i- na- je \textbf{-pai/-ou} na- f\hspace{1.5em}[cop]\label{ex:interrog-co-en}\\
    \textsf{{\sc foc}-} \textsf{I-} \textsf{{\sc fut}-} \textsf{say} \textbf{\textsf{-it/-what}} \textsf{to-} \textsf{him}\\
    `I shall say \textbf{it} to him.' / \\
    `\textbf{What} shall I say to him?' 
    (\begin{coptic}e\0i\0na\0;e\0ou\end{coptic} \begin{coptic}na\0f\end{coptic}) 

\noindent 
The notion of a WH-interrogative construction is a shared level of abstraction that underlies the differences between the languages:
both languages have conventionalized morphosyntactic means to convey that a piece of information is being sought.

Meaning-bearing grammatical constructions such as interrogatives, conditionals, and resultatives are an object of study within and across languages, and many of these have been the focus of semantic/pragmatic annotation schemes, usually involving manual annotation (\cref{sec:identifying}). Our goal is to annotate them on a large scale across many languages in UD treebanks as automatically and accurately as possible. In this paper, we demonstrate how UD treebanks can be enriched with a layer identifying these larger constructions in a typologically informed way so as to enable crosslinguistic comparisons and typological studies.
We present a case study of five construction families and ten languages to illustrate the challenges and opportunities of this approach.

Our goal is challenging because holistic constructions are often not reflected in syntactic labels used in treebanks, which aim to break sentences down into minimal grammatical parts. The UD framework, for example, annotates the individual components of a construction (like the object relation and the interrogative pronoun in \cref{ex:interrog-co-en}) but not the larger whole: there is no `interrogative clause' label in UD. There are other challenges as well. For example, there are many non-canonical and elliptical ways of asking questions in English (e.g., \textit{Can you tell us where?})\ and some questions look identical to exclamations, e.g., \textit{What stunning views}. 

Continuing with the example of interrogative constructions highlights some of the challenges, even within English.
\cref{ex:interrogative-challenges} illustrates 
ambiguity with exclamatives, as well as noncanonical kinds of interrogatives involving ellipsis, idioms, and echo questions.

\ex.\label{ex:interrogative-challenges}
   \a.\label{ex:exclamative} WOW what stunning views. \hfill [en-EWT]\\
(Inferred interpretation: `What stunning views!', not `What stunning views?')
   \b. Can you tell us where. \hfill [en-EWT]
   \c. WELL GUESS WHAT!!! \hfill [en-EWT]
   \d. She didn't have what? \hfill [en-GUM]
   \z.

Thus, defining constructions (or families of related constructions) in crosslinguistically comparable ways, determining what is within scope for annotation in a particular language, and reckoning with ambiguity are all significant challenges.

Despite these challenges, we see constructional annotation as a \textit{worthy} mission for the multilingual computational linguistics community, because the empirical work will deepen understanding of constructional phenomena across languages and provide data for further typological studies.
It is, in our view, also a \textit{viable} way forward,
because the work will draw on the rich ecosystem of UD treebanks and tools in order to add and refine constructional descriptions over time. 
In addition to offering fuller grammatical descriptions of the treebanked sentences, 
construction annotations may be used to improve 
the intra- and interlingual consistency of UD guidelines and data.
On the more practical side, construction annotation could be used for downstream tasks like inducing frame-semantic representations, information extraction, or predicting grammatical difficulty for L2 learners depending on strategies found in the L1 language, or for heritage learners depending on strategies found in the dominant language \citep{bha20}.

To compare across languages, it is necessary to identify patterns larger than a single word or grammatical relation, and to do so in a way that is sensitive to different \emph{morphosyntactic strategies} exhibited by different languages \citep{croft-16,croft2022morphosyntax}.
Our proposed framework, \textbf{UCxn}, is grounded in ideas from Construction Grammar and linguistic typology (\cref{sec:background}).
Our empirical methodology (\cref{sec:methodology}) is to annotate treebanks in each of 10 languages---English, German, Swedish, French, Spanish, Portuguese, Hindi, Mandarin, Hebrew, and Coptic
---for selected constructions by constructing 
graph pattern queries and matching them against UD trees. 
The constructions are interrogatives (\cref{sec:interrogative}), existentials (\cref{sec:existential}), conditionals (\cref{sec:conditional}), resultatives (\cref{sec:resultative}), and noun-adposition-noun combinations where the noun is repeated (NPN; \cref{sec:npn}).
Highlights from our corpus investigations 
corresponding to each construction are discussed in each section, with a quantitative and qualitative discussion in \cref{sec:takeaways}.
\footnote{\label{url}Technical specification, queries and annotated corpora available at \href{https://github.com/LeonieWeissweiler/UCxn}{github.com/LeonieWeissweiler/UCxn}.}

\begin{table*}[h]
\centering
\begin{tabular}{lll}
\toprule
\textbf{Language} & \textbf{Instance}  & \textbf{Query}
\\ \midrule
German &
\raisebox{-.5\height}{\includegraphics[height=1.2cm]{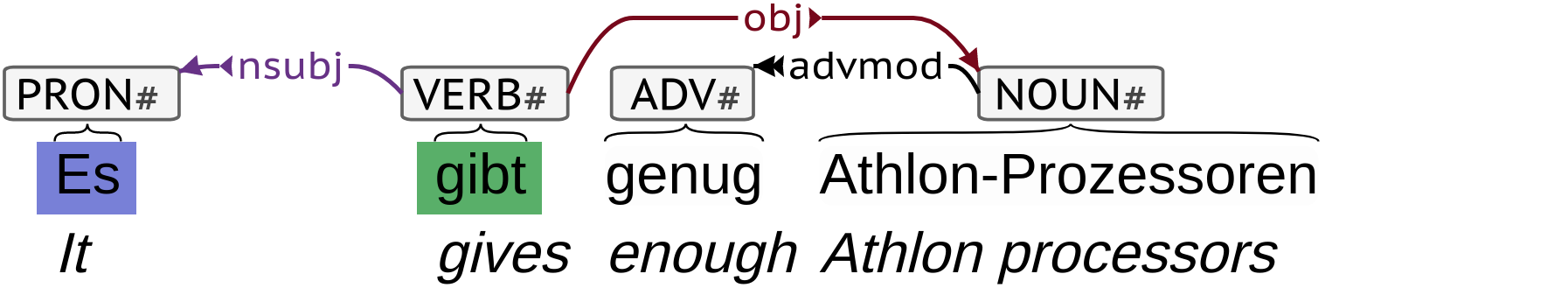}} &
\begin{lstlisting}[language=grew, escapeinside={\%*}{*)}]
pattern
  EXP[lemma=%*``es''*)];
  PRED[lemma=%*``geben''*)];
  PRED-[nsubj]->EXP;
\end{lstlisting}
\\ \midrule
Hebrew &
\raisebox{-.5\height}{\includegraphics[height=1.2cm]{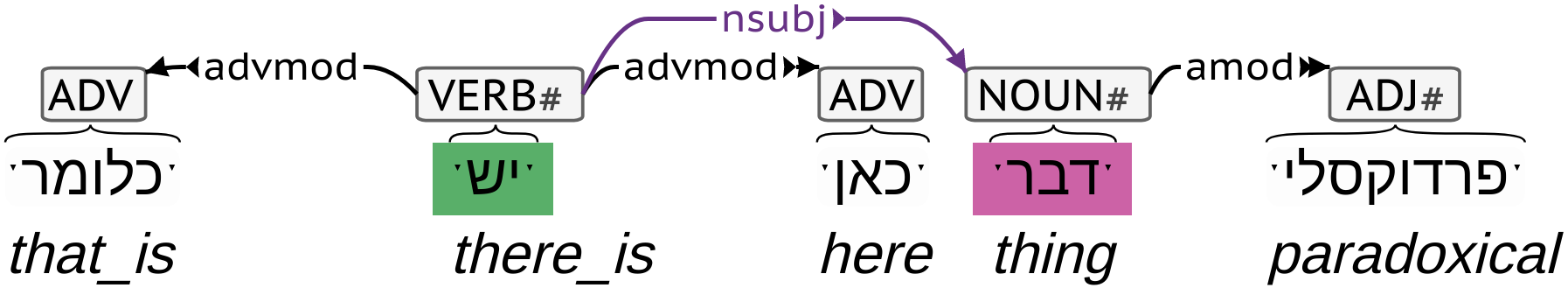}} &
\begin{lstlisting}[language=grew, escapeinside={\%*}{*)}]
pattern
  PRED[lemma=%*``\<y/s>''*)];
  PRED-[nsubj]->PIV;
without
  LE[lemma=%*``\<l>''*)];
  PRED-[obl]->N; N-[case]->LE;
\end{lstlisting}
\\ \midrule
Mandarin &
\raisebox{-.5\height}{\includegraphics[height=1.2cm]{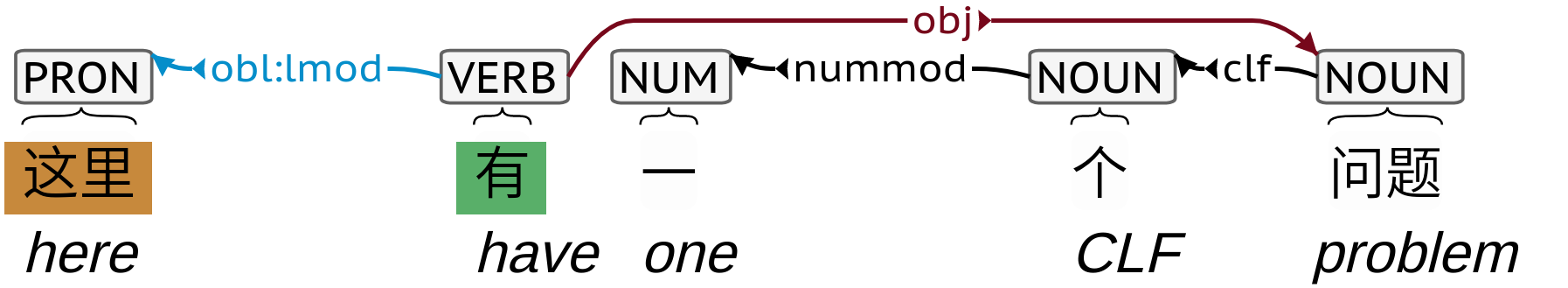}} &
\begin{lstlisting}[language=grew, escapeinside={\%*}{*)}]
pattern
  PRED[form=%*``\begin{CJK}{UTF8}{gbsn}有\end{CJK}''*)];
  PRED-[obl:lmod]->COD;
\end{lstlisting}
\\ \midrule
Spanish &
\raisebox{-.5\height}{\includegraphics[height=1.2cm]{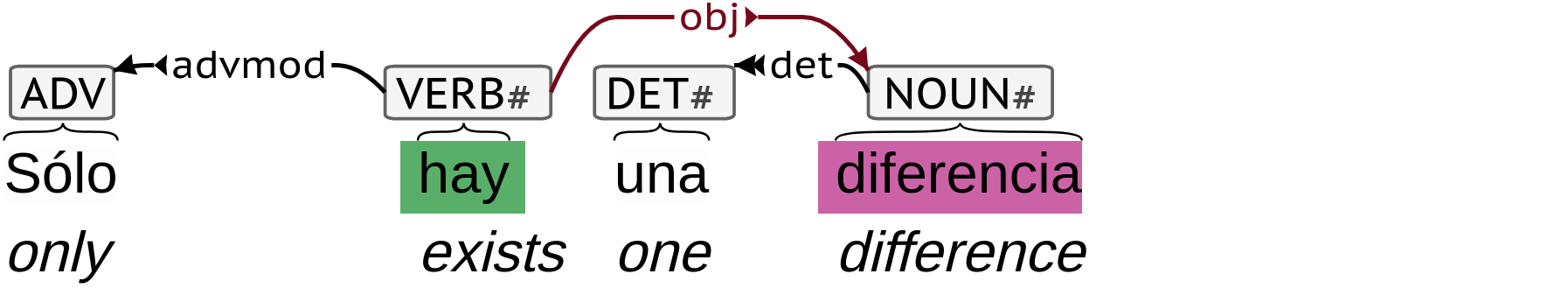}} &
\begin{lstlisting}[language=grew, escapeinside={\%*}{*)}]
pattern
  PRED[lemma=%*``haber''*)];
  PRED-[obj]->PIV;
  DET[upos=DET, Definite=Ind];
  PIV-[det]->DET;
\end{lstlisting} 
 \\
\bottomrule
\end{tabular}
\caption{Existential/presententional construction instances in selected languages and the Grew queries used to identify them. The predicate (\texttt{PRED}), pivot (\texttt{PIV}), coda (\texttt{COD}) and expletive subject (\texttt{EXP}) construction elements and the \deprel{nsubj}, \deprel{obj} and \deprel{obl:lmod} dependency relations are color-coded in the trees and queries.}

\label{tab:grew-match}
\end{table*}

\section{Background}\label{sec:background}

\textbf{Universal Dependencies} (UD) is a framework for crosslinguistically consistent morphosyntactic annotation, which to date has been applied to over 140 languages \citep{nivre16lrec,nivre20lrec,de_marneffe-21}. 
UD annotation consists of two layers: a morphological layer, where each word is assigned a lemma, a universal part-of-speech tag, and a set of morphological features; and a syntactic layer, where all words are connected into a dependency tree labeled with universal syntactic relations. The syntactic representations are defined to prioritize direct relations between content words, which are most likely to be parallel across different languages; function words are treated as grammatical markers on content words. While the inventories of part-of-speech tags and syntactic relations are fixed to support crosslinguistic comparison, the framework allows elaboration through the use of language-specific morphological features and subtypes of syntactic relations.
CoNLL-U is the standard file format for UD treebanks; trees are encoded in 10 tab-separated columns. The last of these, \textsc{misc}, is open-ended to support annotations beyond the UD standard itself.

\paragraph{Construction Grammar}\hspace{-2ex} (CxG) is an approach to linguistic analysis in which the basic unit is a pairing of form and meaning and in which meaning-bearing units can have multiple parts (Construction Elements) \citep{fillmore2012framenet, croft2001radical, fillmore1988regularity, goldberg1995constructions, goldberg2006constructions, hoffmann2013oxford}. A construction grammar is generally represented as a network indicating taxonomic, partonomic and other relations between constructions, called a Constructicon \citep{diessel2019grammar, fillmore2012framenet, lyngfelt2018constructicography}.

\paragraph{Typology} Much work in CxG is done in a single language, where, for example, a specific construction like the English Interrogative Construction is defined by a specific form in English and its functions. This is a \emph{semasiological} approach: it starts from a form and examines its possible functions. However, form varies greatly across languages, and many features of morphosyntactic form are defined in language-specific terms such as specific morphemes (English \lex{do}) or word classes (English Auxiliary). In linguistic typology, one must find a basis for crosslinguistic comparison of a construction, that is, a comparative concept \cite{haspelmath2010comparative}. 

This basis is primarily the function of a construction. For example, a typology of interrogative constructions compares sentence forms across languages that express the function of a speech act requesting information from an interlocutor. This is an \emph{onomasiological} approach: it starts from a function and examines its possible forms. In typology, a construction such as the interrogative construction is the set of form-function pairings across languages expressing a particular function. 
\footnote{In order to distinguish language-specific constructions from constructions as comparative concepts, we follow typological practice and capitalize the names of language-specific constructions.}\footnote{Some work in CxG such as \citet{HasegawaEtAL-comparatives} is onomasiological and crosslinguistic, using frame semantics as the meaning.}

Morphemes and word order can also be described in a crosslinguistically valid fashion. For example, many languages use a special morpheme in interrogative constructions, as with Chinese \begin{CJK*}{UTF8}{gbsn}吗 \end{CJK*} \lex{ma}.
We can describe this as an ``interrogative marker''. 
Other languages change word order in comparison to the declarative construction, albeit in  different ways. 
An interrogative marker or a word order change are two different \emph{strategies} for expressing the interrogative function. 
We can then describe languages as using the same strategy, or different strategies, for the interrogative construction.

Another important concept is \textit{morphosyntactic recruitment}. If two different constructions such as an existential construction and a possessive construction are morphosyntactically similar, we may say that one construction has \emph{recruited} a strategy from the diachronically or conceptually prior construction, although the directionality or their etymological source may not always be clear. 

\paragraph{Related Work}
Prior work on creating datasets annotated with constructions has been in the form of various Constructicon projects, repositories describing the constructions of a language. One of the first and best known is the Berkeley FrameNet Constructicon for English \citep{fillmore2012framenet}. Some Constructicons incorporate UD annotations and corpora \citep[for German, Brazilian Portuguese, and Russian;][]{Ziem.2019,torrent2018,russian_constructicon}. 
While those Constructicons may select individual attestations from corpora to exemplify a construction, in this paper we are concerned with labeling as many instances of the construction in the corpus as possible. Here we take a fundamentally crosslinguistic view of constructions, though the annotation layer could just as well include language-specific constructions. Ultimately we foresee a healthy feedback loop between Constructicon development and corpus enrichment of the kind pursued in this paper.

Construction Grammar has also recently gained popularity in NLP. There have been practical studies using CxG to probe the inner workings of large language models \citep{weissweiler-etal-2022-better, mahowald-2023-discerning}, as well as general observations about the compatibility of usage-based constructionist theories with the recent successes of language models \citep{goldberg2023chat, weissweiler-etal-2023-construction}. Earlier work~\cite{dunn2017,dunietz-etal-2017-corpus, dunietz-etal-2018-deepcx, hwang-palmer-2015-identification} in construction-based NLP focused on the annotation, automatic detection and induction of constructions.

\section{Methodology}\label{sec:methodology}

\paragraph{Selection of Constructions}\label{sec:sel-cxs}

For the purpose of crosslinguistic comparison, we define constructions in terms of function (e.g., a speech act requesting information), rather than form (e.g., subject-auxiliary inversion). We take a modified onomasiological approach: start from a function, and identify the most conventional forms that express the function. In many cases, a language conventionally uses more than one strategy for a construction’s function. We annotated a few, but not all, of the more conventionalized strategies for each construction in each language. Our aim is to see if morphosyntactic queries can detect each strategy in each language, starting only with the information available in UD. 

We chose our constructions to be as diverse as possible. We have selected a speech act construction (interrogative), an information structure construction (existential), a complex sentence construction (conditional), an argument structure construction (resultative), and a phrasal construction (NPN). These constructions cover a broad range of specificity, probably annotation complexity, and size. 
With the NPN construction \citep{Jackendoff_2008}, we examine a strategy, to compare the functions it expresses in the languages in our sample whereas with the other constructions, we examine the functions to compare the strategies recruited in the languages in our sample.

Previous work has explored the relationship between UD annotation and the annotation of (semantically idiosyncratic) multiword expressions \citep{savary-23}. 
Here, by contrast, we focus on constructions that are not fully lexically specified---but we share the goal of identifying structures with more to them semantically than meets the eye.

\paragraph{Selection of Languages}

We select 1 or 2 treebanks for each of a set of languages, ensuring diversity with respect to treebank size and language family.
Each language is worked on by at least one linguist who is also a native or proficient-level speaker. Our languages and the treebanks we use can be seen in \cref{tab:treebank_citations} in \cref{sec:treebanks}.
We use UD~v2.13.

Although our sample of languages is not representative of global language diversity, covering several languages from several regions ensures that we will
cover some variation in strategies.

\paragraph{Identifying Constructions}\label{sec:identifying}

Constructions are defined crosslinguistically in terms of their \emph{function},
but UD annotates morphosyntactic \emph{form}. For some languages and datasets, we do have functional annotations in addition to syntax trees: e.g., the UD English GUM corpus is also annotated with Rhetorical Structure Theory (RST, \citealt{MannThompson1988}), which identifies pragmatic functions for clauses (such as conditional ones), regardless of how they are expressed. Although we can use this type of information to help identify the scope of ways of expressing a certain meaning or class of meanings in a language, we assume that such annotations are either unavailable for most languages, or do not cover the full breadth of functions whose corresponding constructions we are interested in. Our hypothesis is that, in many cases, we can search for the morphosyntactic \emph{strategies} associated with a construction using UD morphosyntactic annotations
and extract tokens of the construction from a treebank with reasonable accuracy.

We test this hypothesis using Grew \citep{guillaume-21}, which allows us to specify search queries with constraints on sentences and their UD annotations, as shown in \cref{tab:grew-match}. For each construction, a language may have multiple Grew patterns corresponding to multiple morphosyntactic strategies. Grew can be combined with Arborator-grew \cite{arborator} to annotate the trees that it finds. 
   
\paragraph{Annotation Atop UD}
We propose a new annotation layer, ``UCxn'', to represent construction instances in UD treebanks.
In our data release, 
UCxn information is incorporated directly into CoNLL-U files, which support arbitrary key-value annotations via the \textsc{misc} field (10th column). 
We introduce the key \texttt{Cxn}, located on the syntactic head token of the construction from the UD tree perspective, i.e.,~the highest-ranking node involved in the construction according to the UD tree, or the earliest such node in case of ties. Construction names are given possibly hierarchical names if subtypes are identifiable, such as \texttt{Interrogative-Polar-Direct} below, to reflect queries at different levels of granularity. 

\begin{table}[h!tb]
\centering
    \resizebox{\columnwidth}{!}{%
\begin{tabular}{llllll}
\toprule
1          & You    & you    & PRON  & ... & \_                            \\
2          & have   & have   & VERB  & ... & Cxn=Interrogative-Polar-Direct \\
3 & a      & a      & DET   & ... & \_                            \\
4 & pencil & pencil & NOUN  & ... & \_                            \\
5 & ?      & ?      & PUNCT & ... & \_             \\
\bottomrule
\end{tabular}
}
\end{table}

\noindent 
A technical specification\footref{url} 
 offers full details on the format and naming conventions in our data.
It also offers the option of annotating \emph{construction elements} in a \texttt{CxnElt} field. 
At present, we annotate only content elements (such as the protasis and the apodosis clauses for conditionals; \cref{sec:conditional}), but not functional elements like subordinators that may be strategy-specific.
Next, we proceed construction by construction, first describing a construction in general terms, 
then highlighting findings from querying treebanks.
\section{Interrogatives}\label{sec:interrogative}

\paragraph{Typological Overview}

An interrogative is a speech act construction, expressing a request for information from the addressee. We focus on clauses realizing two major subfunctions: polarity (``Yes/No'') questions such as \pex{Is she coming?} and information (content, ``WH'') questions such as \emph{Who did you see?}. 
The most common strategies are special prosody, a question marker (see \cref{sec:background}) and special verb forms; less common is a change of word order, as in the English examples above. Content questions contain interrogative phrases such as \pex{who}, \pex{what} or \pex{which (cat)}; their position varies across languages.

\paragraph{Automatic Annotation Efforts}

In this section we compare information questions in which the interrogative phrase is placed either in the same position as its non-interrogative counterpart as in \cref{ex:insitu} or in a different, often fronted position as in \cref{ex:exsitu}.

\ex. You went where? \label{ex:insitu} 

\ex. Where did you go? \label{ex:exsitu}

To identify interrogatives, we relied on either the presence of WH items (\lex{what}, \lex{who}, etc.), word order (in languages using it for marking), as well as the presence of question marks or sentence type annotations where available. In some languages, WH items are identical to indefinite pronouns or free-relative heads (e.g.,~\pex{I ate what you cooked} is not interrogative, despite containing \lex{what}), but the UD morphological feature \texttt{PronType=Int} helps to disambiguate.
We did not see the special verb form strategy in our treebanks.

\Cref{tab:eng-situ} shows pre- and post-posed realization frequencies for different grammatical functions for WH pronouns in interrogatives (i.e.,~excluding uses such as `I know who!'), compared to overall usage excluding such pronouns. The table shows the strong preference to front WH objects in English (28:3 in favor of pre-posed; for other objects the ratio is 265:8889). For other functions, the picture is more complex: interrogative adverbials such as `when' and `where' appear almost exclusively preposed, while non-interrogative phrases strongly prefer fronting, but only at a rate of 8258/2196 (79\%). 

\begin{table}[t]\centering\small
\begin{tabularx}{\columnwidth}{Xlrrrr}
\toprule
     & & \multicolumn{2}{c}{\textbf{Non-interrog.}} & \multicolumn{2}{c}{\textbf{Interrog.}} \\
\cmidrule(lr){3-4}\cmidrule(lr){5-6}
& & \multicolumn{1}{c}{\textbf{pre}}          & \multicolumn{1}{c}{\textbf{post}}          & \multicolumn{1}{c}{\textbf{pre}}        & \textbf{post}        \\ \hline
\multirow{8}{*}{\shortstack{\textbf{English}\\\textbf{(GUM)}}}&\textit{advmod} & 8258                  & 2196                   & 122                 & 1                    \\
&\textit{nsubj}  & 14512                 & 500                    & 50                  & 0                    \\
&\textit{obj}    & 265                   & 8889                   & 28                  & 3                    \\
&\textit{det}    & 15985                 & 36                     & 26                  & 0                    \\
&\textit{obl}    & 1255                  & 7867                   & 6                   & 1                    \\
&\textit{ccomp}  & 142                   & 1370                   & 4                   & 0                    \\
&\textit{xcomp}  & 15                    & 2831                   & 4                   & 0                    \\
&\textit{other}  & 139                   & 8732                   & 4                   & 1                   \\
\midrule
\multirow{5}{*}{\textbf{Coptic}}&\textit{advmod} & 1110                  & 1702                   & 1                   & 3             \\
&\textit{nsubj}  & 4844                  & 575                    & 5                   & 2                    \\
&\textit{obj}    & 2                     & 2585                   & 0                   & 15                   \\
&\textit{obl}    & 228                   & 4339                   & 35                  & 23                   \\
&\textit{ccomp}  & 0                     & 750                    & 0                   & 43                   \\
&\textit{other}  & 2                     & 2478                   & 2                   & 15                \\ \bottomrule  
\end{tabularx}
\caption{Pre- and post-posed dependent WH pronouns and non-WH equivalents in EN and COP.}
\label{tab:cop-situ}
\label{tab:eng-situ}
\end{table}

Turning to Coptic for comparison, \cref{tab:cop-situ} shows a rather different picture.
The tendency for placing subjects before their heads and objects after them is much weaker (5:2, but based on only 7 cases); for adverbial interrogatives, fronting occurs proportionally less than in non-interrogatives, though there is very little data. 
The frequent presence of the Coptic focalizing marker \lex{ere}, which indicates a contrast with a previously uttered or implied phrase, plays a role in promoting late realization of arguments, above and beyond the tendency for each grammatical function \cite{GreenReintges2001}. 

\paragraph{Takeaways} Although typological literature often classifies languages in terms of basic word order or the possibility of word order changes, 
the actual picture in individual language data is much more complex. 
We have shown that quantitative analyses with construction-annotated data give a more nuanced picture of how languages realize such word order dependencies in interrogatives. 

\section{Existentials}\label{sec:existential}

\paragraph{Typological Overview}

Existentials assert the existence (or not) of an entity (`pivot'), almost always indefinite, and usually specified in a location (`coda'), as in \pex{There are yaks in Tibet}. This function is closely related to the presentational function, introducing a referent, as in \pex{There's a yak on the road}. 
%In our language sample, the two functions are expressed using the same strategies, such that their distinction may be largely context-dependent. For this reason, we consider here both existentials and presentatives.
As the two functions are often formally indistinguishable, especially when taken out of context, we consider here both existentials and presentatives.

Languages vary with respect to the predicate that they use in the existential. 
One class of languages employ a construction-specific lexeme, such as Swedish \lex{finnas}. 
In Coptic there are lexicalized negative and positive existence predicates,  \begin{coptic}oun\end{coptic} \lex{oun} and \begin{coptic}mmn\end{coptic} \lex{mmn}. 
Historically, predicative possession used the same items, but through lexicalization, the possessive versions are now lexically distinct from the existentials.

The relationship between existence and possession also has synchronic manifestations.
Our sample includes languages that use a possession verb as the predicate in an existential,
one predicate to express both existence and possession,
such as \lex{ter} `to have' in Brazilian Portuguese, French \lex{avoir} in the phrase \pex{il y a}, or 
the Mandarin predicate \begin{CJK*}{UTF8}{gbsn}有\end{CJK*} \lex{yǒu}. This duality is also found in Hebrew, where possession is expressed by adding a possessive dative argument to the existential construction \cref{ex:HEB-HAYU1}.

\exg. \textsf{hayu} \textsf{(la-nu)} \textsf{kama} \textsf{taxanot} \textsf{ba-derex}\\
    \textsf{were.3{\sc p}} \textsf{(to-us)} \textsf{few} \textsf{stops.{\sc pf}} \textsf{in.the-way}  \\
    `There were/We had a few stops on the way.' \\ (\<bdrK> \<t.hnwt> \<kmh> (\<lnw>) \<hyw>) \hfill [he-HTB]  \label{ex:HEB-HAYU1}

An additional existential strategy shares a copula with the predicational locative construction. In Hebrew, the copula \lex{\<hyh>} \lex{haya} is used in past and future tense existentials (\pex{hayu} in \cref{ex:HEB-HAYU1} is the inflected form). For Mandarin, the use of the copula \begin{CJK*}{UTF8}{gbsn}是\end{CJK*}  \lex{shì} is an alternative to the lexicalized existential predicate \begin{CJK*}{UTF8}{gbsn}有\end{CJK*}  \lex{yǒu}. The link between locative and existential
is also found in locatives that grammaticalized into unique existential forms such as English \lex{There('s)} or French \lex{y} (in \pex{il y a}). 

Finally, the existential predicates \lex{haber} and \lex{haver} in Spanish and Portuguese, respectively, also function as auxiliaries and modals, similarly to the English \lex{have}, modulo possession.

The argument structure of existential predicates is not uniform crosslinguistically, with pivots exhibiting different degrees of subjecthood properties \citep{Keenan.1976}. This diversity is manifested in the UD annotation. 
In one class of languages, no argument is identified as \deprel{nsubj} and the pivot is attached as \deprel{obj} in UD. This is the case in Spanish and Mandarin (see \cref{tab:grew-match}). 

Other languages identify the pivot as \deprel{nsubj}. This is the case in Hebrew, where the copula standardly exhibits agreement with the pivot, as in \cref{ex:HEB-HAYU1}. However, unlike typical subjects, the Hebrew pivot appears post-verbally, does not always trigger agreement, and in informal speech may receive accusative marking, if definite. Likewise, in Coptic the pivot is \deprel{nsubj} in postverbal position,
though the adverb \lex{there} is added in around 5\% of cases in the UD data with no clear antecedent.

A different strategy involves employing an expletive as a co-argument to the pivot. This is found in our language sample in French and English \cref{ex:EXPL-FR} and in German (\cref{tab:grew-match}).

\exg. \textsf{il} \textsf{y} \textsf{a} \textsf{une} \textsf{salle} \textsf{à l'étage} \\
 \textsf{it} \textsf{there} \textsf{has} \textsf{a} \textsf{room} \textsf{upstairs}  \\
    `There is a room upstairs.' \hfill [fr-GSD]\label{ex:EXPL-FR}

Here, too, UD annotations vary across languages. In the English treebank the pivot is attached as \deprel{nsubj} and \lex{there} as \deprel{expl}. In French, the expletive \lex{y} is \deprel{expl:comp} and the pivot is \deprel{obj}. In German, the expletive \lex{es} is \deprel{nsubj} and the pivot is \deprel{obj}.

\paragraph{Automatic Annotation Efforts}
Our languages vary in the difficulty of identifying existential constructions. The easiest cases were those in which a construction-specific lexical item is employed (e.g., the lexicalized existential predicates in Coptic and Swedish). In French, instances of the existentials are identified by queries which target the construction-specific cooccurrence of the clitic \lex{y} and the verb \lex{avoir} in a \deprel{comp:expl} relation.

The more challenging cases are those in which the elements which encode existence are multi-functional. In some treebanks, this challenge is overcome by construction-specific annotations. Thus, for example, in the Hebrew HTB the predicate \lex{\<hyh>} \lex{haya} is annotated as \texttt{HebExistential=Yes} where it is used in its existential function. 
% \az{but then it is not a copula, right? So maybe better 'the predicate'?}. \nm{We use the term `copula' in the paragraph which refers to example \ref{ex:HEB-HAYU}. `Predicate' is too general, because we want to distinguish between special predicates, copulas and \emph{have} predicates.} 

%\sh{Nurit, I changed and erased some parts. Is it ok?}\nm{yes}
When disambiguating annotations are not available, the queries rely on other distributional properties of the construction to avoid false positives. In French, the queries only target indefinite pivots, excluding definite determiners and numerals. 
%Similarly, in Spanish, we target the lemma \lex{haber} with an indefinite complement.  \nm{What are "NPs that have determinative indefinite pronouns"? What do you mean by "we can add" -- Do our queries retrieve such cases? }  
In Hebrew, the queries exclude instances where the predicate has a \deprel{obl} dependent with a dative case marker, i.e., a possessor (see query in Table \ref{tab:grew-match}). Furthermore, to distinguish between the predicational and existential functions of  \lex{\<hyh>} \lex{haya} in UD\_Hebrew-IAHLTwiki \cite{zeldes-etal-2022-second}, where this information is not annotated, the query targets only post-verbal \deprel{nsubj} dependents. 

\paragraph{Takeaways} Lexical items that are associated with the existential construction are often shared with other constructions. For this reason, in order to maximize accuracy the queries cannot only rely on these lexical items but also target morphosyntactic properties and dependency relations.

\section{Conditionals}\label{sec:conditional}

\paragraph{Typological Overview}

A conditional construction is a complex sentence construction describing a broadly ``causal'' link between the two states of affairs, the protasis (condition) and the apodosis (consequence) \citep[pp.~81--82]{comrie1986}.
The strategies for conditional constructions are largely the typical ones for complex sentences in general \citep[pp.~532--34]{croft2022morphosyntax}. The construction may be an adverbial subordinate construction or a coordinate construction. The clauses may be balanced (identical in form to a declarative main clause) or deranked (one clause, usually the protasis, is in a distinct form, with a special verb form and other differences).
There may be a subordinating conjunction such as \lex{if}, or rarely a change in word order, as in English \pex{Had he stayed, he would have seen it.} The nonfactual nature of conditionals may manifest in irrealis or subjunctive verb forms.

\paragraph{Automatic Annotation Efforts}

Common strategies for conditionals are the use of a subordinating conjunction as in German \textit{Wenn die Möglichkeit da ist} (lit.~`if the opportunity there is') 
(3291 instances in German-HDT, 240 in Swedish-Talbanken, 243 in Hindi-HDT, 495 in English-GUM), or word order inversion as in Swedish \textit{Har du god kondition} (lit. have you good condition; if you are in good shape)
(1182 instances in German-HDT, 68 in Swedish-Talbanken, 7 in English-GUM). A very different strategy involves conditional circumfixes. 
In Coptic \lex{e- -šan} is a circumfix that conveys conditionality (CD) and applies to the pronominal subject of the conditional clause so that \textit{e-f-šan-eibe} (CD-he-CD-thirst) means \textit{If he is thirsty.}

Our investigation of conditionals has shown that it may not be possible, using the information available in UD, to create queries that accurately retrieve conditional sentences.
There are three sources of difficulty: (1)~the need for information that is not yet encoded in UD, (2)~subordinating conjunctions and clause types that are not exclusively used in conditional constructions, and (3)~the variety of subordinating conjunctions and other strategies that are used to express the conditional construction.

Conditional subordinating conjunctions can be divided into: simple subordinating conjunctions like \lex{agar}/\lex{yadi} (Hindi) (`if'); complex subordinating conjunctions like \lex{förutsatt att} (Swedish) (`provided that'); and V2 sentence embedders like \lex{angenommen} (German) (`presumed') \cite{Breindl.2014}. Complex subordinating conjunctions and V2 sentence embedders are problematic in German HDT because the part of speech is not \textit{conjunction} and the dependency label is not \textit{mark}  (or \textit{fixed expression} as in Swedish). The query needs to specify the connector lemma, giving many false positives. 

In Germanic languages, conditional constructions without subordinating conjunctions usually express the protasis as verb-initial clauses that precede the main clause. In Swedish and German, any verb can be used in a verb-initial protasis clause. In English, however, it is restricted to certain auxiliaries
(e.g.~\pex{\textbf{Had} I gone, I would have seen you}).

While the subtypes of English conditionals (e.g., neutral or negative epistemic stance) require many search queries but are in principle findable with UD, this is not the case in German.
A major problem for German conditionals---especially with regard to semantic and syntactic subcategorization---was posed by the inadequate mood annotations. German HDT does not annotate conditional or potential verb forms and marks most verb forms as indicative, even when there is a clear conditional or subjunctive structure. It is therefore not possible to search for semantic subcategories based on different mood annotations in HDT, although verb mood is the most common indication of grouping conditionals in German \cite{Schierholz.2022b}. 

\paragraph{Takeaways}

The conditional strategies are in principle searchable, although writing these rules requires an exhaustive study of the phenomenon in each language.
Search requirements vary in complexity depending on the depth of the underlying linguistic analyses of the phenomenon. Annotation practices may complicate the search process and even make some distinctions impossible. 

\section{Resultatives}\label{sec:resultative}

\paragraph{Typological Overview}
From a functional perspective the resultative construction expresses an event with two subevents: a \emph{dynamic} subevent such as \lex{paint} and a \emph{resulting state} subevent such as \lex{red} in \cref{ex:RES}.

\ex.\label{ex:RES}\label{RES1} They painted the door \textbf{red}. 

The English resultative construction is a prime example of an argument structure construction \citep{hovav2001event, goldberg2004english}. A basic transitive clause describing an event is augmented with a secondary predicate describing the result state of a participant, but there are many strategies to express this function. English, for example, also uses adverbial subordination of the dynamic event: \pex{The door was red as a result of their painting it} or \pex{I flattened the metal by hammering}. In our study of the resultative construction, we are only annotating cases where the language provides a conventionalized strategy for expressing the resultative event as a complex predicate composed of a dynamic action and a result state.

\paragraph{Automatic Annotation Efforts} In the sample languages, we encountered several challenges. First, in some languages a resultative conceptualization is lacking: they do not combine a dynamic event with a stative result event into a complex predicate. In Hebrew, the most natural way of expressing the painting event  
literally translates as `They painted the door in red' (the result expressed with an oblique marked by the prepositional prefix \lex{be-} ‘in’). 
In Hindi, complex predicates expressing a cause-result relation 
have a dynamic event as a result \cref{ex:hi-resultative}. 
We consider these languages as lacking the resultative construction as defined above. 
\exg. ... ki veh duSman ko \textbf{maar} \textbf{bhagaa-ye} \hspace{8.4em}[hi-HUTB]\\ 
... that it enemy {\sc acc} \textbf{hit} \textbf{run.{\sc caus-subj}}\\
`... that it beat and chase away the enemy.'\\[-.5ex] (\raisebox{-0.36\height}{\includegraphics[height=19pt]{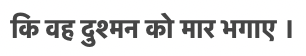}})
\label{ex:hi-resultative}

Second, in several languages many of the complex events with a dynamic subevent and resulting state subevent were of the form [`make/do' X STATE], where the dynamic event is the causative verb `make/do', e.g., Hindi \lex{kar} `do', Swedish \lex{göra} `make/do', or German \lex{machen} as in \textit{Nvidia machts möglich} `Nvidia makes it possible'. This construction is generally analyzed as the causative of a stative event, and is excluded from the resultative category. In the German and Swedish treebanks, removal of the causative left few or no examples of genuine resultatives.

Third, in some languages, the UD annotation of the resultative construction is indistinguishable from another construction such as depictive secondary predication. For example, \pex{I hammered the metal flat} (resultative) has the same structure as \pex{I left the door open} (depictive).
It would be necessary in English for queries to incorporate lexical lists of predicates licensing the construction, in order to disambiguate from other sentences with similar UD structures, at the expense of generalizability to predicates that have not been seen in the resultative construction. 

Finally, Chinese has a very productive resultative construction \cref{ex:RES-MAN1}, which is already annotated in the treebank Chinese-HK \cite{ud_hk_2017} with a label specifically designed for resultative complements: \texttt{compound:vv}. They are trivially extracted by querying for that dependency relation.

\exg.
   wǒ \textbf{qiāo} \textbf{píng} le dīngzi\hspace{3.5em}[zh-HK]\label{ex:RES-MAN1}\label{ex:RES-MAN1a}\\
    1{\sc sg} \textbf{hit} \textbf{flat} {\sc perf} nail\\
    `I hammered the nail flat.' 
    (\begin{CJK*}{UTF8}{gbsn}我敲平了钉子.\end{CJK*})

\paragraph{Takeaways}
In summary, the attempt to annotate the resultative construction has shown us several difficulties: annotating a construction where boundaries are in dispute within the literature, which might not even exist in all languages depending on the definition, and where considerable linguistic expertise and manual effort is required to write a comprehensive set of rules, indicating the need for collaboration among theoretical linguists, corpus linguists, typologists and computational linguists.  Efforts such as ours can reveal constructions that need further linguistic investigation, or can help solidify linguistic consensus on the definition of the construction. 

\begin{table}[t]
\centering\small
\begin{tabularx}{\columnwidth}{lYYYYY}
\toprule
\textbf{Lang.} & \textbf{SU} & \textbf{CO} & \textbf{OP} & \textbf{PR} & \textbf{QU}        \\ \midrule

COP & + & \textminus & + & \textminus & (+)         \\        
EN & + & + & + & + & +                  \\
FR& + & (+) & + & + & (+)                  \\
DE & + & \textminus & + & + & +                  \\
HE & + & + & + & + & (+)                   \\

HI & (?) & (?) & (?) & \textminus & \textminus                  \\
ZH & (?) & \textminus & \textminus & \textminus & \textminus                  \\ 
PT & + & + & + & + & (+)              \\
ES & + & + & + & + & (+)                  \\
SV & + & (+) & (+) & + & +           \\
\bottomrule
\end{tabularx}
\caption{Semantic categories of NPN and their crosslinguistic attestation in UD treebanks. \textminus~means that the target meaning is not possible in the language. (+)~signals that the meaning is possible but not attested in the UD treebanks. (?) means that the existence of this meaning is unclear, see footnote \ref{foot:npn}. Succession: SU, Comparison: CO, Opposition: OP, Proximity: PR, Quantification: QU}
\label{tab:npn}
\end{table}

\begin{table*}
    \centering\small
    \setlength{\tabcolsep}{4pt}
    \begin{tabular}{@{}>{\bf}lcccccrr@{}}
        \toprule
        Lang.   
        &  Interrogative (\cref{sec:interrogative}) 
        & Existential (\cref{sec:existential}) 
        &  Conditional (\cref{sec:conditional}) 
        & Resultative (\cref{sec:resultative})
        & NPN (\cref{sec:npn}) 
        & total sent. & total tokens \\
         \midrule
         EN 
         &  1117; 769
         & 472; 319 (\texttt{f})
         & 762; 375 (\texttt{D})
         & \texttt{H}, \texttt{D}
         & 21; 12 
         & 17k; 11k & 254k; 187k\\
         DE 
         &  5483 (\texttt{H})
         & 3392 (\texttt{H})
         & 3291 (\texttt{A},\texttt{H}) 
         & \texttt{D}
         & 40  
         & 190k & 3.5m\\
         SV 
         &  \hphantom{0}276\hphantom{ (X)}
         & \hphantom{0}235\hphantom{ (X)} 
         & \hphantom{0}310 (\texttt{H})\hphantom{,X}
         & \texttt{D}
         & \hphantom{0}7 
         & \hphantom{00}6k & \hphantom{0}96k\\
         FR 
         &   \hphantom{0}368\hphantom{ (X)}
         & \hphantom{0}114 (\texttt{F})
         & \hphantom{0}213 (\texttt{F})\hphantom{,X}
         & \texttt{D}
         & 12  
         & \hphantom{0}16k & 400k\\
         ES 
         &  \hphantom{0}580\hphantom{ (X)}
         & \hphantom{0}160 (\texttt{F})
         & \hphantom{0}502 (\texttt{F})\hphantom{,X}
         & \texttt{D}
         & 37  
         & \hphantom{0}18k & 567k\\
         PT 
         &  \hphantom{0}337 (\texttt{A}) 
         & \hphantom{0}340 (\texttt{F})
         &  \hphantom{0}106\hphantom{ (X,X)}
         & \texttt{D}
         & \hphantom{0}7  
         & \hphantom{00}9k &227k\\
         HI  
         & \hphantom{0}285\hphantom{ (X)}
         & 2058 (\texttt{F}) 
         & \hphantom{0}350 (\texttt{A})\hphantom{,X}
         & \texttt{D}
         & \hphantom{0}\texttt{?} 
         & \hphantom{0}16k & 351k \\
         ZH
         &  \hphantom{0}146\hphantom{ (X)}
         &  \hphantom{00}58 (\texttt{F}) 
         &  \hphantom{00}31\hphantom{ (X,X)}
         &  78 (\texttt{D}) 
         & \hphantom{0}\texttt{?} 
         & \hphantom{00}1k & \hphantom{00}9k\\
         HE  
         &  236; 22 
         &  113; 60
         &  192; 56 
         & \texttt{D}
         & 9; 11 
         & 6k; 5k & 160k; 140k\\
         COP  
         & \hphantom{0}150\hphantom{ (X)}
         & \hphantom{00}80\hphantom{ (X)} 
         & \hphantom{0}185\hphantom{ (X,X)}
         & \texttt{D}
         & \hphantom{0}2 
         & \hphantom{00}2k & \hphantom{0}55k\\     
         \bottomrule
    \end{tabular}
    \caption{Counts of identified construction instances by treebank, along with qualifications: definitional issues (\texttt{D}), UD annotation errors (\texttt{A}), occasional false positives (\texttt{f}), frequent false positives (\texttt{F}), 
    unattested strategies (\texttt{H}). \texttt{?} means that the existence of the productive construction is doubtful (see \cref{foot:npn}). The two numbers for EN and HE represent the two treebanks for each (see~\cref{tab:treebank_citations} in the Appendix).}
    \label{tab:issues}

\end{table*}

\section{NPN}\label{sec:npn}

\paragraph{Typological Overview}

With the preceding four constructions, we took an onomasiological approach, examining them crosslinguistically on a functional basis.
Most work in CxG, however, takes a semasiological (form-first) approach to characterizing a formal pattern and its function(s), usually within a single language. In our terms, this approach starts with a strategy and examines the range of functions using that strategy.
The UD framework offers a common vocabulary for describing formal categories of morphology, parts of speech, and grammatical relations across languages.
In this section, we consider how a semasiological or strategy-based inquiry can be conducted crosslinguistically using UD corpora.
As a case study, we look at the ``NPN'' strategy, in which a meaning related to quantification or iteration is expressed with a repeated noun and an adposition or case marker on the second noun. 
Examples in English include \pex{day after day, shoulder to shoulder, box upon box,} etc.
While infrequent and often a source of idioms, this strategy recurs across many languages \citep{Postma_1995,Matsuyama_2004,Jackendoff_2008,König_Moyse-Faurie_2009,Roch_Keßelmeier_Muller_2010,Pskit_2015,Pskit_2017,Kinn_2022}.\footnote{The  studies cover Dutch, English, French, German, Norwegian, Japanese, Mandarin, Polish, and Spanish.}

\paragraph{Automatic Annotation Efforts}

We find examples of NPN strategies across 8/10 languages.\footnote{We did not find any attestations of NPN in the Chinese or Hindi treebanks. It is unclear whether NPN is productive in these languages, but we are aware of expressions that might qualify: e.g.,~Mandarin \lex{yī tiān bǐ yī tiān} and Hindi \lex{din ba din} (both `day by day').\label{foot:npn}}
In our queries, we limit ourselves to instances where the two Ns are the same lemma, though there are related NPN uses where the two Ns are not the same \citep{Jackendoff_2008}.   A few examples of NPN from our treebanks are presented in \cref{ex:pt-npn}.

\ex. PT: \lex{frente a frente} `face to face' (lit.~`front to front'), FR: \lex{jour pour jour} `to the day' (lit.~`day for day'), SV: \lex{steg för steg} `step by step' (lit.~`step for step'), HE: \lex{mila be-mila} `word for word' (lit.~`word in word')
\label{ex:pt-npn} 

In terms of morphosyntactic form, NPN strategies are well captured by our queries because of the strict precedence relationship between the constituent elements. We find that there is considerable variability in whether NPNs are analyzed as fixed expressions in UD (using the \deprel{fixed} relation type), or whether the second N is analyzed as an \deprel{nmod} of the first N.
\footnote{We restrict our queries to exclude cases where the first N is marked by another adposition, because we find that in many languages the PNPN strategy (\pex{from time to time}) has a different range of meanings than the NPN strategy.
We also exclude cases where nouns are modified with adjectives, as these are extremely rare.}

The semantics of NPN have been well investigated in previous literature \citep{Jackendoff_2008,Roch_Keßelmeier_Muller_2010,Sommerer_Baumann_2021,Kinn_2022}. We find that most of the previously proposed semantic subcategories emerge in our languages. Following the categorization and discussion in \citet{Jackendoff_2008} and later works, we find the following semantic subtypes of NPN: \textsc{succession} (\pex{hour after hour}), \textsc{comparison} (\pex{man for man}), \textsc{opposition} (\pex{brother against brother}), \textsc{proximity} (\pex{hand in hand}) and \textsc{quantification} (particularly of a large quantity, \pex{snacks upon snacks}). 
Qualitatively, we noticed that the \textsc{succession} submeaning was most prevalent% 
%across our treebanks
, and \textsc{opposition} is typically restricted to body parts, as in \cref{ex:pt-npn}. 
\Cref{tab:npn} summarizes our empirical findings by semantic subtype and language.

\paragraph{Takeaways}
Using a strategy, like NPN, as the basis of typological comparison is not without issue \citep{croft2022morphosyntax}; however, we do find considerable functional overlap in terms of the meanings which are conveyed by the NPN strategy in our language sample. Notably, NPN is the only investigated construction/strategy for which the query is almost universal across languages, meaning that it is the most well-integrated with UD: if the promise is universality across languages, then ideally a query would also work across all languages. It makes perfect sense that this only works with strategies, which are defined by their form, and not for constructions, which are defined by their meaning, as UD itself focuses on form.

\section{Survey Summary}\label{sec:takeaways}

Our 5~case studies have surveyed constructions and strategies in 10~languages.
\Cref{tab:issues} provides a quantitative summary in terms of matched instances per treebank.
Treebanks ranged in size from 9k to 3.5m tokens; in some cases, the scale was too small for a robust set of results.
NPN was particularly sparse---this is simply a rare strategy.

\Cref{tab:issues} also provides a qualitative summary of some of the major kinds of issues encountered:
definitional issues (\texttt{D}), 
annotation errors in the treebank (\texttt{A}), 
unavoidable occasional false positives (\texttt{f}),
many false positives due to overlap with another construction (\texttt{F}),
and unattested strategies for which at least one query returned 0 examples (\texttt{H}).
For most of the languages, we abandoned attempts to quantify resultatives given the definitional challenges.
Note that some of the larger treebanks had unattested strategies (\texttt{H})---this is not necessarily because of a problem with the treebanks, but reflects that more effort was put into writing queries for long-tail strategies in those languages.

We are pleased to see that UD annotation errors (\texttt{A}) were not a major source of difficulty for most of the treebanks examined. 
On the other hand, many constructions were fundamentally difficult to circumscribe (\texttt{D}) or distinguish from other constructions given the available UD annotations (\texttt{F}).
These may necessitate human annotation and/or supplementary information from semantic analyzers.

For English and Hebrew, where we consulted two treebanks, we can see some differences in the construction counts that are not explained by the size of the treebank but rather by the domain. 
This underscores the importance of domain diversity in empirical studies of constructions.

\section{Conclusion and Future Work}

We have presented a case study of annotating constructions in UD treebanks. We developed automatic annotation queries for ten languages and five construction families, and 
developed UCxn as a framework for representing them in UD treebanks. Overall, we find that annotating constructions is feasible with a mix of automatic and manual efforts, and that with typologically-based construction definitions, the annotations support crosslinguistic quantitative studies.

The next step is to scale up our approach to more languages and constructions, possibly with the aid of construction parsers (and/or UD parsers to produce larger-scale silver treebanks for investigating rare constructions). Beyond the created resources, these efforts may prompt improvements to the UD annotation guidelines and to language-specific Constructicons.
Crucially, this work has been a first attempt at bringing two important frameworks together. We aim to gather feedback and input from the community to further our goal of integrating constructions fully with UD. 

\section*{Acknowledgments}
This work was initiated by the Dagstuhl Seminar 23191 ``Universals of Linguistic Idiosyncrasy in Multilingual Computational Linguistics'' (\url{https://www.dagstuhl.de/23191}). In addition to the authors, our discussion group also included Francis Bond, Jorg Bücker, Mathieu Constant, Daniel Flickinger, Sylvain Kahane, Peter Ljunglöf, Teresa Lynn, Alexandre Rademaker, Manfred Sailer, and Agata Savary. We are grateful for Grew infrastructure support from Bruno Guillaume; for feedback from members of the NERT lab at Georgetown and anonymous reviewers; and for discussion with Natália Sathler Sigliano about some of the constructions in Portuguese. 
This work was supported in part by Israeli Ministry of Science and Technology grant No.~0002336 (Nurit Melnik, PI), CAPES PROEX grant No.~88887.816228/2023-00 (Arthur Lorenzi, PhD) and NSF award IIS-2144881 (Nathan Schneider, PI).

\section*{Bibliographical References}\label{sec:reference}

\bibliographystyle{lrec_natbib}
\bibliography{unicodex,anthology,Ninas-references,languageresource}

\begin{thebibliography}{61}
\expandafter\ifx\csname natexlab\endcsname\relax\def\natexlab#1{#1}\fi

\bibitem[{Bast et~al.(2021)Bast, Endresen, Janda, Lund, Lyashevskaya, McDonald,
  Mordashova, Nesset, Rakhilina, Tyers, and Zhukova}]{russian_constructicon}
Radovan Bast, Anna Endresen, Laura~A. Janda, Marianne Lund, Olga Lyashevskaya,
  James McDonald, Daria Mordashova, Tore Nesset, Ekaterina Rakhilina,
  Francis~M. Tyers, and Valentina Zhukova. 2021.
\newblock \href {https://constructicon.github.io/russian/} {The {R}ussian
  {C}onstructicon. {A}n electronic database of the {R}ussian grammatical
  constructions}.
\newblock Available at \url{https://constructicon.github.io/russian/}.

\bibitem[{Bhat et~al.(2017)Bhat, Bhatt, Farudi, Klassen, Narasimhan, Palmer,
  Rambow, Sharma, Vaidya, Vishnu et~al.}]{treebank_hi}
Riyaz~Ahmad Bhat, Rajesh Bhatt, Annahita Farudi, Prescott Klassen, Bhuvana
  Narasimhan, Martha Palmer, Owen Rambow, Dipti~Misra Sharma, Ashwini Vaidya,
  Sri~Ramagurumurthy Vishnu, et~al. 2017.
\newblock The {H}indi/{U}rdu treebank project.
\newblock In \emph{Handbook of Linguistic Annotation}, pages 659--697. Springer
  Press.

\bibitem[{Bhatia and Montrul(2020)}]{bha20}
Archna Bhatia and Silvina Montrul. 2020.
\newblock Comprehension of differential object marking by {H}indi heritage
  speakers.
\newblock In A.~Mardale and S.~Montrul, editors, \emph{The Acquisition of
  Differential Object Marking}, pages 261--281. John Benjamins Publishing
  Company.

\bibitem[{Borges~V{\"o}lker et~al.(2019)Borges~V{\"o}lker, Wendt, Hennig, and
  K{\"o}hn}]{borges-volker-etal-2019-hdt}
Emanuel Borges~V{\"o}lker, Maximilian Wendt, Felix Hennig, and Arne K{\"o}hn.
  2019.
\newblock \href {https://doi.org/10.18653/v1/W19-8006} {{HDT}-{UD}: A very
  large {U}niversal {D}ependencies treebank for {G}erman}.
\newblock In \emph{Proceedings of the Third Workshop on Universal Dependencies
  (UDW, SyntaxFest 2019)}, pages 46--57, Paris, France. Association for
  Computational Linguistics.

\bibitem[{Breindl et~al.(2014)Breindl, Volodina, and Wa{\ss}ner}]{Breindl.2014}
Eva Breindl, Anna Volodina, and Ulrich~Hermann Wa{\ss}ner. 2014.
\newblock \href {https://doi.org/10.1515/9783110341447} {\emph{Handbuch der
  deutschen Konnektoren 2: Semantik der deutschen Satzverkn{\"u}pfer}}, volume
  Band 13 of \emph{Schriften des Instituts f{\"u}r Deutsche Sprache}.
\newblock {De Gruyter}, Berlin and M{\"u}nchen and Boston.

\bibitem[{Comrie(1986)}]{comrie1986}
Bernard Comrie. 1986.
\newblock Conditionals: a typology.
\newblock In E.~C. Traugott, A.~ter Meulen, J.~S. Reilly, and C.~A. Ferguson,
  editors, \emph{On Conditionals}, pages 77--99. Cambridge University Press.

\bibitem[{Croft(2001)}]{croft2001radical}
William Croft. 2001.
\newblock \emph{Radical {C}onstruction {G}rammar: Syntactic theory in
  typological perspective}.
\newblock Oxford University Press, Oxford, {UK}.

\bibitem[{Croft(2016)}]{croft-16}
William Croft. 2016.
\newblock \href {https://doi.org/10.1515/lingty-2016-0012} {Comparative
  concepts and language-specific categories: {T}heory and practice}.
\newblock \emph{Linguistic Typology}, 20(2):377--393.
\newblock Publisher: De Gruyter Mouton.

\bibitem[{Croft(2022)}]{croft2022morphosyntax}
William Croft. 2022.
\newblock \href {https://doi.org/10.1017/9781316145289} {\emph{Morphosyntax:
  {C}onstructions of the world's languages}}.
\newblock Cambridge University Press.

\bibitem[{de~Marneffe et~al.(2021)de~Marneffe, Manning, Nivre, and
  Zeman}]{de_marneffe-21}
{Marie-Catherine} de~Marneffe, Christopher~D. Manning, Joakim Nivre, and Daniel
  Zeman. 2021.
\newblock \href {https://doi.org/10.1162/coli_a_00402} {Universal
  {D}ependencies}.
\newblock \emph{Computational Linguistics}, 47(2):255--308.

\bibitem[{Diessel(2019)}]{diessel2019grammar}
Holger Diessel. 2019.
\newblock \emph{The Grammar Network: How Linguistic Structure is Shaped by
  Language Use}.
\newblock Cambridge University Press, Cambridge.

\bibitem[{Dunietz et~al.(2018)Dunietz, Carbonell, and
  Levin}]{dunietz-etal-2018-deepcx}
Jesse Dunietz, Jaime Carbonell, and Lori Levin. 2018.
\newblock \href {https://doi.org/10.18653/v1/D18-1196} {{D}eep{C}x: A
  transition-based approach for shallow semantic parsing with complex
  constructional triggers}.
\newblock In \emph{Proceedings of the 2018 Conference on Empirical Methods in
  Natural Language Processing}, pages 1691--1701, Brussels, Belgium.
  Association for Computational Linguistics.

\bibitem[{Dunietz et~al.(2017)Dunietz, Levin, and
  Carbonell}]{dunietz-etal-2017-corpus}
Jesse Dunietz, Lori Levin, and Jaime Carbonell. 2017.
\newblock \href {https://doi.org/10.18653/v1/W17-0812} {The {BEC}au{SE} corpus
  2.0: Annotating causality and overlapping relations}.
\newblock In \emph{Proceedings of the 11th Linguistic Annotation Workshop},
  pages 95--104, Valencia, Spain. Association for Computational Linguistics.

\bibitem[{Dunn(2017)}]{dunn2017}
Jonathan Dunn. 2017.
\newblock Computational learning of construction grammars.
\newblock \emph{Language and cognition}, 9(2):254--292.

\bibitem[{Einarsson(1976)}]{talbanken_1}
Jan Einarsson. 1976.
\newblock \emph{Talbankens skriftspr{\aa}kskonkordans}.
\newblock Institutionen f{\"o}r nordiska spr{\aa}k, Lunds universitet.

\bibitem[{Fillmore et~al.(1988)Fillmore, Kay, and
  O'Connor}]{fillmore1988regularity}
Charles~J. Fillmore, Paul Kay, and Mary~Catherine O'Connor. 1988.
\newblock \href {http://www.jstor.org/stable/414531} {Regularity and
  idiomaticity in grammatical constructions: the case of `let alone'}.
\newblock \emph{Language}, 64(3):501--538.

\bibitem[{Fillmore et~al.(2012)Fillmore, {Lee-Goldman}, and
  Rhodes}]{fillmore2012framenet}
Charles~J. Fillmore, Russell~R. {Lee-Goldman}, and Russell Rhodes. 2012.
\newblock \href
  {http://loven.gu.se/infoglueCalendar/digitalAssets/1775658128_BifogadFil_Framenetconstructicon.pdf}
  {The {FrameNet} {C}onstructicon}.
\newblock In Hans~C. Boas and Ivan~A. Sag, editors, \emph{{Sign-Based}
  Construction Grammar}, pages 283--322. {CSLI} Publications, Stanford, {CA}.

\bibitem[{Goldberg(1995)}]{goldberg1995constructions}
Adele~E. Goldberg. 1995.
\newblock \emph{Constructions: a construction grammar approach to argument
  structure}.
\newblock University of Chicago Press, Chicago.

\bibitem[{Goldberg(2006)}]{goldberg2006constructions}
Adele~E. Goldberg. 2006.
\newblock \emph{Constructions at Work: The Nature of Generalization in
  Language}.
\newblock Oxford University Press, Oxford.

\bibitem[{Goldberg(To appear)}]{goldberg2023chat}
Adele~E. Goldberg. To appear.
\newblock A chat about constructionist approaches and {LLMs}.
\newblock \emph{Constructions and Frames}.

\bibitem[{Goldberg and Jackendoff(2004)}]{goldberg2004english}
Adele~E Goldberg and Ray Jackendoff. 2004.
\newblock The english resultative as a family of constructions.
\newblock \emph{Language}, 80(3):532--568.

\bibitem[{Green and Reintges(2001)}]{GreenReintges2001}
Melanie Green and Chris~H. Reintges. 2001.
\newblock Syntactic anchoring in {H}ausa and {C}optic wh-constructions.
\newblock \emph{Proceedings of the Annual Meeting of the Berkeley Linguistics
  Society}, 27(2).

\bibitem[{Guibon et~al.(2020)Guibon, Courtin, Gerdes, and
  Guillaume}]{arborator}
Gaël Guibon, Marine Courtin, Kim Gerdes, and Bruno Guillaume. 2020.
\newblock \href {https://www.aclweb.org/anthology/2020.lrec-1.651} {When
  collaborative treebank curation meets graph grammars}.
\newblock In \emph{Proceedings of The 12th Language Resources and Evaluation
  Conference}, pages 5293--5302, Marseille, France. European Language Resources
  Association.

\bibitem[{Guillaume(2021)}]{guillaume-21}
Bruno Guillaume. 2021.
\newblock \href {https://doi.org/10.18653/v1/2021.eacl-demos.21} {Graph
  matching and graph rewriting: {GREW} tools for corpus exploration,
  maintenance and conversion}.
\newblock In \emph{Proceedings of the 16th Conference of the European Chapter
  of the Association for Computational Linguistics: System Demonstrations},
  pages 168--175, Online. Association for Computational Linguistics.

\bibitem[{Guillaume et~al.(2019)Guillaume, de~Marneffe, and
  Perrier}]{guillaume-etal-2019-conversion}
Bruno Guillaume, Marie-Catherine de~Marneffe, and Guy Perrier. 2019.
\newblock \href {https://aclanthology.org/2019.tal-2.4} {Conversion et
  am{\'e}liorations de corpus du fran{\c{c}}ais annot{\'e}s en {U}niversal
  {D}ependencies [conversion and improvement of {U}niversal {D}ependencies
  {F}rench corpora]}.
\newblock \emph{Traitement Automatique des Langues}, 60(2):71--95.

\bibitem[{Hasegawa et~al.(2010)Hasegawa, {Lee-Goldman}, Ohara, Fujii, and
  Fillmore}]{HasegawaEtAL-comparatives}
Yoko Hasegawa, Russell {Lee-Goldman}, Kyoko~Hirose Ohara, Seiko Fujii, and
  Charles~J. Fillmore. 2010.
\newblock On expressing measurement and comparison in {E}nglish and {J}apanese.
\newblock In Hans~C. Boas, editor, \emph{Contrastive Studies in Construction
  Grammar}, pages 169--200. John Benjamins, Amsterdam.

\bibitem[{Haspelmath(2010)}]{haspelmath2010comparative}
Martin Haspelmath. 2010.
\newblock Comparative concepts and descriptive categories in crosslinguistic
  studies.
\newblock \emph{Language}, 86(3):663--687.

\bibitem[{Hoffmann and Trousdale(2013)}]{hoffmann2013oxford}
Thomas Hoffmann and Graeme Trousdale. 2013.
\newblock \emph{The Oxford handbook of construction grammar}.
\newblock Oxford University Press.

\bibitem[{Hovav and Levin(2001)}]{hovav2001event}
Malka~Rappaport Hovav and Beth Levin. 2001.
\newblock An event structure account of english resultatives.
\newblock \emph{Language}, 77(4):766--797.

\bibitem[{Hwang and Palmer(2015)}]{hwang-palmer-2015-identification}
Jena~D. Hwang and Martha Palmer. 2015.
\newblock \href {https://doi.org/10.18653/v1/S15-1006} {Identification of
  caused motion construction}.
\newblock In \emph{Proceedings of the Fourth Joint Conference on Lexical and
  Computational Semantics}, pages 51--60, Denver, Colorado. Association for
  Computational Linguistics.

\bibitem[{Jackendoff(2008)}]{Jackendoff_2008}
Ray Jackendoff. 2008.
\newblock “{C}onstruction after {C}onstruction” and {I}ts {T}heoretical
  {C}hallenges.
\newblock \emph{Language}, 84(1):8–28.

\bibitem[{Keenan(1976)}]{Keenan.1976}
Edward Keenan. 1976.
\newblock Towards a universal definition of subject.
\newblock In Charles~N. Li, editor, \emph{Subject and Topic}, pages 303--334.
  Academic Press New York, New York.

\bibitem[{Kinn(2022)}]{Kinn_2022}
Torodd Kinn. 2022.
\newblock \href {https://doi.org/10.1017/S0022226721000116} {Regular and
  compositional aspects of {NPN} constructions}.
\newblock \emph{Journal of Linguistics}, 58(1):1–35.

\bibitem[{König and Moyse-Faurie(2009)}]{König_Moyse-Faurie_2009}
Ekkehard König and Claire Moyse-Faurie. 2009.
\newblock \href {https://doi.org/10.1515/9783110216134.1.57} {{S}patial
  reciprocity: between grammar and lexis}.
\newblock In \emph{Form and Function in Language Research: Papers in Honour of
  Christian Lehmann}, page 57–68. De Gruyter Mouton.

\bibitem[{Lyngfelt et~al.(2018)Lyngfelt, Borin, Ohara, and
  Torrent}]{lyngfelt2018constructicography}
Benjamin Lyngfelt, Lars Borin, Kyoko Ohara, and Tiago~Timponi Torrent. 2018.
\newblock \emph{Constructicography: Constructicon development across
  languages}, volume~22.
\newblock John Benjamins Publishing Company.

\bibitem[{Mahowald(2023)}]{mahowald-2023-discerning}
Kyle Mahowald. 2023.
\newblock \href {https://aclanthology.org/2023.eacl-main.20} {A discerning
  several thousand judgments: {GPT}-3 rates the article + adjective + numeral +
  noun construction}.
\newblock In \emph{Proceedings of the 17th Conference of the European Chapter
  of the Association for Computational Linguistics}, pages 265--273, Dubrovnik,
  Croatia. Association for Computational Linguistics.

\bibitem[{Mann and Thompson(1988)}]{MannThompson1988}
William~C. Mann and Sandra~A. Thompson. 1988.
\newblock {Rhetorical Structure Theory}: Toward a functional theory of text
  organization.
\newblock \emph{Text}, 8(3):243--281.

\bibitem[{Matsuyama(2004)}]{Matsuyama_2004}
Tetsuya Matsuyama. 2004.
\newblock \href {https://doi.org/10.9793/elsj1984.21.55} {{T}he {N} {A}fter {N}
  {C}onstruction}.
\newblock \emph{English Linguistics}, 21(1):55–84.

\bibitem[{Nivre et~al.(2016)Nivre, de~Marneffe, Ginter, Goldberg, Haji\v{c},
  Manning, McDonald, Petrov, Pyysalo, Silveira, Tsarfaty, and
  Zeman}]{nivre16lrec}
Joakim Nivre, Marie-Catherine de~Marneffe, Filip Ginter, Yoav Goldberg, Jan
  Haji\v{c}, Christopher~D. Manning, Ryan McDonald, Slav Petrov, Sampo Pyysalo,
  Natalia Silveira, Reut Tsarfaty, and Dan Zeman. 2016.
\newblock {U}niversal {D}ependencies v1: {A} multilingual treebank collection.
\newblock In \emph{Proceedings of the 10th International Conference on Language
  Resources and Evaluation ({LREC})}, pages 1659--1666.

\bibitem[{Nivre et~al.(2020)Nivre, de~Marneffe, Ginter, Haji\v{c}, Manning,
  Pyysalo, Schuster, Tyers, and Zeman}]{nivre20lrec}
Joakim Nivre, Marie-Catherine de~Marneffe, Filip Ginter, Jan Haji\v{c},
  Christopher~D. Manning, Sampo Pyysalo, Sebastian Schuster, Francis Tyers, and
  Dan Zeman. 2020.
\newblock {U}niversal {D}ependencies v2: {An} evergrowing multilingual treebank
  collection.
\newblock In \emph{Proceedings of the 12th International Conference on Language
  Resources and Evaluation ({LREC})}, pages 4034--4043.

\bibitem[{Nivre et~al.(2006)Nivre, Nilsson, and
  Hall}]{nivre-etal-2006-talbanken05}
Joakim Nivre, Jens Nilsson, and Johan Hall. 2006.
\newblock \href {http://www.lrec-conf.org/proceedings/lrec2006/pdf/223_pdf.pdf}
  {{T}albanken05: A {S}wedish treebank with phrase structure and dependency
  annotation}.
\newblock In \emph{Proceedings of the Fifth International Conference on
  Language Resources and Evaluation ({LREC}{'}06)}, Genoa, Italy. European
  Language Resources Association (ELRA).

\bibitem[{Postma(1995)}]{Postma_1995}
Gertjan Postma. 1995.
\newblock \href {https://doi.org/10.1075/avt.12.17pos} {{Z}ero {S}emantics —
  {T}he syntactic encoding of quantificational meaning}.
\newblock \emph{Linguistics in the Netherlands}, 12:175–190.

\bibitem[{Pskit(2015)}]{Pskit_2015}
Wiktor Pskit. 2015.
\newblock {T}he {C}ategorial {S}tatus and {I}nternal {S}tructure of {NPN}
  {F}orms in {E}nglish.
\newblock In \emph{{W}ithin {L}anguage, {B}eyond {T}heories (Volume I):
  {S}tudies in {T}heoretical {L}inguistics}, page 27–42. Cambridge Scholars
  Publishing.

\bibitem[{Pskit(2017)}]{Pskit_2017}
Wiktor Pskit. 2017.
\newblock \href {https://www.ceeol.com/search/chapter-detail?id=648266}
  {Linguistic and philosophical approaches to {NPN} structures}.
\newblock In \emph{Topics in Syntax and Semantics. Linguistic and Philosophical
  Perspectives}, page 93–110. Wydawnictwo Uniwersytetu.

\bibitem[{Rademaker et~al.(2017)Rademaker, Chalub, Real, Freitas, Bick, and
  de~Paiva}]{rademaker-etal-2017-universal}
Alexandre Rademaker, Fabricio Chalub, Livy Real, Cl{\'a}udia Freitas, Eckhard
  Bick, and Valeria de~Paiva. 2017.
\newblock \href {https://aclanthology.org/W17-6523} {{U}niversal {D}ependencies
  for {P}ortuguese}.
\newblock In \emph{Proceedings of the Fourth International Conference on
  Dependency Linguistics (Depling 2017)}, pages 197--206, Pisa,Italy.
  Link{\"o}ping University Electronic Press.

\bibitem[{Roch et~al.(2010)Roch, Keßelmeier, and
  Muller}]{Roch_Keßelmeier_Muller_2010}
Claudia Roch, Katja Keßelmeier, and Antje Muller. 2010.
\newblock Productivity of {NPN} sequences in {G}erman, {E}nglish, {F}rench, and
  {S}panish.
\newblock In \emph{Proceedings of the Conference on Natural Language Processing
  2010}, page 158–163, Saarbrücken, Germany.

\bibitem[{Sade et~al.(2018)Sade, Seker, and Tsarfaty}]{sade-etal-2018-hebrew}
Shoval Sade, Amit Seker, and Reut Tsarfaty. 2018.
\newblock \href {https://doi.org/10.18653/v1/W18-6016} {The {H}ebrew
  {U}niversal {D}ependency treebank: Past present and future}.
\newblock In \emph{Proceedings of the Second Workshop on Universal Dependencies
  ({UDW} 2018)}, pages 133--143, Brussels, Belgium. Association for
  Computational Linguistics.

\bibitem[{Sanguinetti et~al.(2022)Sanguinetti, Cassidy, Bosco, \"{O}zlem
  \c{C}etino\u{g}lu, Cignarella, Lynn, Rehbein, Ruppenhofer, Seddah, and
  Zeldes}]{SanguinettiEtAl2022}
Manuela Sanguinetti, Lauren Cassidy, Cristina Bosco, \"{O}zlem
  \c{C}etino\u{g}lu, Alessandra~Teresa Cignarella, Teresa Lynn, Ines Rehbein,
  Josef Ruppenhofer, Djam\'{e} Seddah, and Amir Zeldes. 2022.
\newblock Treebanking user-generated content: a ud based overview of
  guidelines, corpora and unified recommendations.
\newblock \emph{Language Resources and Evaluation}, 57:493--544.

\bibitem[{Savary et~al.(2023)Savary, Stymne, Mititelu, Schneider, Ramisch, and
  Nivre}]{savary-23}
Agata Savary, Sara Stymne, Verginica~Barbu Mititelu, Nathan Schneider, Carlos
  Ramisch, and Joakim Nivre. 2023.
\newblock \href {https://nejlt.ep.liu.se/article/view/4453} {{PARSEME} {M}eets
  {U}niversal {D}ependencies: {G}etting on the same page in representing
  multiword expressions}.
\newblock \emph{Northern European Journal of Language Technology}, 9(1).

\bibitem[{Schierholz and Uzonyi(2022)}]{Schierholz.2022b}
Stefan~J. Schierholz and P{\'a}l Uzonyi, editors. 2022.
\newblock \href {https://doi.org/10.1515/9783110698527} {\emph{Grammatik: Band
  2: Syntax}}, volume Bd. 1.2 of \emph{W{\"o}rterb{\"u}cher zur Sprach- und
  Kommunikationswissenschaft}.
\newblock {De Gruyter}, Berlin and Boston.

\bibitem[{Silveira et~al.(2014)Silveira, Dozat, de~Marneffe, Bowman, Connor,
  Bauer, and Manning}]{silveira-etal-2014-gold}
Natalia Silveira, Timothy Dozat, Marie-Catherine de~Marneffe, Samuel Bowman,
  Miriam Connor, John Bauer, and Chris Manning. 2014.
\newblock \href
  {http://www.lrec-conf.org/proceedings/lrec2014/pdf/1089_Paper.pdf} {A gold
  standard dependency corpus for {E}nglish}.
\newblock In \emph{Proceedings of the Ninth International Conference on
  Language Resources and Evaluation ({LREC}'14)}, pages 2897--2904, Reykjavik,
  Iceland. European Language Resources Association (ELRA).

\bibitem[{Sommerer and Baumann(2021)}]{Sommerer_Baumann_2021}
Lotte Sommerer and Andreas Baumann. 2021.
\newblock \href {https://doi.org/10.1515/cog-2020-0013} {Of absent mothers,
  strong sisters and peculiar daughters: The constructional network of english
  {NPN} constructions}.
\newblock \emph{Cognitive Linguistics}, 32(1):97–131.

\bibitem[{Taul{\'e} et~al.(2008)Taul{\'e}, Mart{\'\i}, and
  Recasens}]{taule-etal-2008-ancora}
Mariona Taul{\'e}, M.~Ant{\`o}nia Mart{\'\i}, and Marta Recasens. 2008.
\newblock \href
  {http://www.lrec-conf.org/proceedings/lrec2008/pdf/35_paper.pdf} {{A}n{C}ora:
  Multilevel annotated corpora for {C}atalan and {S}panish}.
\newblock In \emph{Proceedings of the Sixth International Conference on
  Language Resources and Evaluation ({LREC}'08)}, Marrakech, Morocco. European
  Language Resources Association (ELRA).

\bibitem[{Torrent et~al.(2018)Torrent, Matos, Lage, Laviola, da~Silva~Tavares,
  de~Almeida, and Sigiliano}]{torrent2018}
Tiago~Timponi Torrent, Ely~Edison Matos, Ludmila~Meireles Lage, Adrieli
  Laviola, Tatiane da~Silva~Tavares, V\^ania~Gomes de~Almeida, and
  Nat\'alia~Sathler Sigiliano. 2018.
\newblock \href {https://doi.org/10.1075/cal.22.04tor} {Towards continuity
  between the lexicon and the constructicon in {FrameNet} {B}rasil}.
\newblock In Benjamin Lyngfelt, Lars Borin, Kyoko Ohara, and Tiago~Timponi
  Torrent, editors, \emph{Constructicography: {C}onstructicon development
  across languages}, pages 107--140. John Benjamins, Amsterdam.

\bibitem[{Weissweiler et~al.(2023)Weissweiler, He, Otani, R.~Mortensen, Levin,
  and Sch{\"u}tze}]{weissweiler-etal-2023-construction}
Leonie Weissweiler, Taiqi He, Naoki Otani, David R.~Mortensen, Lori Levin, and
  Hinrich Sch{\"u}tze. 2023.
\newblock \href {https://aclanthology.org/2023.cxgsnlp-1.10} {Construction
  grammar provides unique insight into neural language models}.
\newblock In \emph{Proceedings of the First International Workshop on
  Construction Grammars and NLP (CxGs+NLP, GURT/SyntaxFest 2023)}, pages
  85--95, Washington, D.C. Association for Computational Linguistics.

\bibitem[{Weissweiler et~al.(2022)Weissweiler, Hofmann, K{\"o}ksal, and
  Sch{\"u}tze}]{weissweiler-etal-2022-better}
Leonie Weissweiler, Valentin Hofmann, Abdullatif K{\"o}ksal, and Hinrich
  Sch{\"u}tze. 2022.
\newblock \href {https://doi.org/10.18653/v1/2022.emnlp-main.746} {The better
  your syntax, the better your semantics? probing pretrained language models
  for the {E}nglish comparative correlative}.
\newblock In \emph{Proceedings of the 2022 Conference on Empirical Methods in
  Natural Language Processing}, pages 10859--10882, Abu Dhabi, United Arab
  Emirates. Association for Computational Linguistics.

\bibitem[{Wong et~al.(2017)Wong, Gerdes, Leung, and Lee}]{ud_hk_2017}
Tak-sum Wong, Kim Gerdes, Herman Leung, and John Lee. 2017.
\newblock \href {https://aclanthology.org/W17-6530} {Quantitative comparative
  syntax on the {C}antonese-{M}andarin parallel dependency treebank}.
\newblock In \emph{Proceedings of the Fourth International Conference on
  Dependency Linguistics (Depling 2017)}, pages 266--275, Pisa, Italy.
  Link{\"o}ping University Electronic Press.

\bibitem[{Zeldes(2017)}]{zeldes-17}
Amir Zeldes. 2017.
\newblock \href {http://dx.doi.org/10.1007/s10579-016-9343-x} {The {GUM}
  corpus: creating multilayer resources in the classroom}.
\newblock \emph{Language Resources and Evaluation}, 51(3):581--612.

\bibitem[{Zeldes and Abrams(2018)}]{zeldes-18}
Amir Zeldes and Mitchell Abrams. 2018.
\newblock \href {https://www.aclweb.org/anthology/W18-6022} {The {C}optic
  {U}niversal {D}ependency {T}reebank}.
\newblock In \emph{Proc. of the Second Workshop on Universal Dependencies
  ({UDW} 2018)}, pages 192--201, Brussels, Belgium.

\bibitem[{Zeldes et~al.(2022)Zeldes, Howell, Ordan, and
  Ben~Moshe}]{zeldes-etal-2022-second}
Amir Zeldes, Nick Howell, Noam Ordan, and Yifat Ben~Moshe. 2022.
\newblock \href {https://doi.org/10.18653/v1/2022.emnlp-main.292} {A second
  wave of {UD} {H}ebrew treebanking and cross-domain parsing}.
\newblock In \emph{Proceedings of the 2022 Conference on Empirical Methods in
  Natural Language Processing}, pages 4331--4344, Abu Dhabi, United Arab
  Emirates. Association for Computational Linguistics.

\bibitem[{Ziem et~al.(2019)Ziem, Flick, and Sandkühler}]{Ziem.2019}
Alexander Ziem, Johanna Flick, and Phillip Sandkühler. 2019.
\newblock \href {https://doi.org/10.1515/lex-2019-0003} {The german
  constructicon project: Framework, methodology, resources}.
\newblock \emph{Lexicographica}, 35(2019):61--86.

\end{thebibliography}

\onecolumn
\appendix

\begin{multicols}{2}

\section{List of Treebanks}\label[appsec]{sec:treebanks}
An overview of the treebanks used in this work along with their total number of sentences is provided in Table \ref{tab:treebank_citations}. The genres covered by each treebank are shown in \cref{tab:treebank_genres}.

\columnbreak

\hphantom{0}

\end{multicols}

\begin{table}[h]
    \centering\small
    \begin{tabular}{@{}c@{~}l@{}rr} 
    \toprule
         \textbf{Lang} & \multicolumn{1}{c}{\textbf{Treebanks}} &Num. Sents \\
         \midrule
         \textbf{EN} & EWT, GUM \textsmaller{\citep{silveira-etal-2014-gold,zeldes-17}} &16,662; 10,761\\
         \textbf{DE} & HDT \textsmaller{\cite{borges-volker-etal-2019-hdt}}&189,928\\
         \textbf{SV} & Talbanken \textsmaller{\citep{talbanken_1, nivre-etal-2006-talbanken05}} &6,026\\
         \textbf{FR} & GSD \textsmaller{\cite{guillaume-etal-2019-conversion}}&16,342\\
         \textbf{ES} & AnCora \textsmaller{\cite{taule-etal-2008-ancora}}&17,662\\
         \textbf{PT} & Bosque \textsmaller{\cite{rademaker-etal-2017-universal}}&9,357\\
         \textbf{HI} & HUTB \textsmaller{\cite{treebank_hi}} &15,649\\
         \textbf{ZH} & Chinese-HK \textsmaller{\cite{ud_hk_2017}}&1,004\\
         \textbf{HE} & HTB, IAHLTwiki \textsmaller{\cite{sade-etal-2018-hebrew,zeldes-etal-2022-second}} & 6,143; 5,039\\

         \textbf{COP} & Coptic Scriptorium \textsmaller{\citep{zeldes-18}} &2,203\\
         \bottomrule
    \end{tabular}
    \caption{UD treebanks used in our crosslinguistic study. Some cover specific varieties, e.g., AnCora represents European Spanish, whereas Bosque covers both European and Brazilian Portuguese. Chinese is limited to Mandarin. Coptic (Sahidic) is the only historical language.}
    \label{tab:treebank_citations}
\end{table}

\begin{table*}[h]
\centering\small
\begin{tabular}{l|cccccccccc}
\toprule
                         & \textbf{EN} & \textbf{DE} & \textbf{SV} & \textbf{FR} & \textbf{ES} & \textbf{PT} & \textbf{HI} & \textbf{ZH} & \textbf{HE} & \textbf{COP} \\
                         \hline
\textbf{academic}        & +           &             &             &             &             &             &             &             &             &             \\
\textbf{bible}           &             &             &             &             &             &             &             &             &             & +           \\
\textbf{blog}            & +           &             &             &       +      &             &             &             &             &             &             \\
\textbf{e-mail}          & +           &             &             &             &             &             &             &             &             &             \\
\textbf{fiction}         & +           &             &             &             &             &             &             &             &             & +           \\
\textbf{government}      & +           &             &             &             &             &             &             &             &             &             \\
\textbf{grammar examples} &             &             &             &             &             &             &             &             &             &             \\
\textbf{learner essays}   &             &             &             &             &             &             &             &             &             &             \\
\textbf{legal}           &             &             &             &             &             &             &             &             &             &             \\
\textbf{medical}         &             &             &             &             &             &             &             &             &             &             \\
\textbf{news}            & +           &     +        &     +        &      +       &       +      & +           &    +         &             &       +      &             \\
\textbf{nonfiction}      & +           &      +       &       +      &             &             &             &             &             &             & +           \\
\textbf{poetry}          &             &             &             &             &             &             &             &             &             &             \\
\textbf{reviews}         & +           &             &             &       +      &             &             &             &             &             &             \\
\textbf{social}          & +           &             &             &             &             &             &             &             &             &             \\
\textbf{spoken}          & +           &             &             &             &             &             &             &      +       &             &             \\
\textbf{web}             & +           &       +      &             &             &             &             &             &             &             &             \\
\textbf{wiki}            & +           &             &             &       +      &             &             &             &             &        +     &            \\
\bottomrule
\end{tabular}
    \caption{Genres covered by the UD treebanks used in the paper. }
    \label{tab:treebank_genres}
\end{table*}

\end{document}